\documentclass[conference]{IEEEtran}
\usepackage{amsmath,amssymb}
\usepackage{graphicx}
\usepackage{booktabs}
\usepackage{color}
\usepackage[hidelinks]{hyperref}

\newif\ifdraftmode \draftmodefalse
\newcommand{\pending}[1]{\ifdraftmode{\color{blue}\scriptsize[\textsc{pending human data:} #1]}\fi}

\begin{document}

\title{Consistency Without Alignment: Item-Sensitive Language Models Indistinguishable From Random}
\author{
\IEEEauthorblockN{Cris Huynh}
\IEEEauthorblockA{Independent researcher\\crishuynh2004@gmail.com}
}
\maketitle

\begin{abstract}
Item-sensitivity, whether a model's choice depends on the specific input rather
than on its own output prior, is widely reported as evidence that a model is
doing a task. We show this evidence is necessary but not sufficient, in a
forced-choice signalling task, abstracted from the board game \emph{Deception:
Murder in Hong Kong}, where the reference points a coordinate should be
judged against, a fit-maximising strategy, a posterior-maximising strategy, and
uniform random selection, are all computable in closed form. Across seven
language models, two model families, a post-training ablation, and three
independent scoring rules, every one of 21 model-by-rule cells is reliably
item-sensitive. Yet 8 of those 21 cells are not statistically distinguishable
from a chooser that ignores the item and selects at random, and 5 score worse
than random at describing the target. Item-sensitivity and distance from random
correlate at only $r=0.30$. We call this consistency without alignment and
argue it generalises to any evaluation that reports item-sensitivity,
permutation consistency, or self-consistency as evidence of task competence
without an independent reference for the quantity being measured. We further
find that a literal-similarity baseline with no pragmatics outperforms most of
the language models tested, that adding a pragmatic layer over two baseline
similarity sources moves choosers toward random rather than toward the
Bayesian reference, and that a standard labelled multiple-choice format carries
no measurable content signal here. All results are on the model side of a
pre-registered instrument; a matched human condition is designed and piloted
but not yet collected.
\end{abstract}

\section{Introduction}
Item-sensitivity metrics, whether a model's output depends on the specific
input it is shown rather than on its own output prior, are widely used as
evidence that a model is doing a task: permutation consistency, self-consistency
across rephrasings, and agreement across resamples all take this form. We show
that this evidence is necessary but not sufficient, in a setting where we can
compute what it is missing exactly.

Seven language models, spanning four orders of magnitude in parameter count,
two model families, and a post-training ablation, are reliably item-sensitive
on the tile we analyse. Every model responds to which item it is shown, under
every one of three token-normalisation rules we test, with zero exceptions
across 21 model-by-rule cells. Yet on the coordinate the task is built to
measure, a chooser's position between a fit-maximising strategy and a
posterior-maximising one, eight of those same 21 cells are not statistically
distinguishable from a chooser that ignores the item and selects uniformly at
random. Five cells score worse than random at describing the target. Across
cells, how item-sensitive a model is correlates with how far its coordinate sits
from random at only $r=0.30$. Knowing that a model responds reliably to an input
tells you little about whether it responds to anything the experimenter cares
about.

We call this \emph{consistency without alignment}, and take it to be a hazard
for any evaluation that reports an item-sensitivity statistic as evidence of
task competence, on any task. The statistic alone has no reference point.
Without a baseline computed independently of the model, a measure of dependence
on the input cannot distinguish ``responds to the feature we intend to
measure'' from ``responds reliably to some other feature of the input.''
Demonstrating the gap requires a setting where that independent reference is
computable, not estimated or assumed, which is what the rest of this paper
builds.

\paragraph{The setting} is a constrained cooperative channel embedded in an
adversarial audience: one party knows a hidden target pair of concepts and must
communicate it by selecting a single option from a menu that was never designed
to express it, forced lossy encoding. It is abstracted from the board game
\emph{Deception: Murder in Hong Kong} (Tobey Ho, Grey Fox Games), in which a
player who alone knows a hidden pair of cards may communicate them to the other
players only by placing markers on a fixed set of categorical scene tiles;
Section~\ref{sec:setting} states what we take from it and what we do not. We
use it because the two competing
accounts of how a chooser resolves that constraint, select the best-fitting
description or select the most discriminating one, have closed-form coordinates
here: the hypothesis space is small enough to enumerate exactly and the
underlying similarity function is a closed-form quantity fit to human judgements
rather than to text or to model output. A third reference, uniform random
selection, is equally closed-form. All three give a model's coordinate something
external to be measured against. The setting is the instrument; the claim above
is what it is for.

\paragraph{Contributions}
\begin{itemize}
\item We show that item-sensitivity and alignment with either of two normative
criteria are empirically dissociable. Excess, our item-sensitivity statistic, is
positive in 21 of 21 model-by-rule cells; the same 21 cells' coordinates range
from indistinguishable from random to 91 percent of the way to a salience
reference, with no relationship strong enough to predict one from the other
($r=0.30$).
\item We give a general diagnosis: item-sensitivity metrics need an independent
reference for the quantity of interest, not only a null for the metric itself.
We build the reference this task was missing, a marginal null for excess and a
random-selection baseline for the coordinate, and show that the second one
changes what an existing headline result means (Section~\ref{sec:align}).
\item We report where the models tested actually land. A mandatory literal
cosine-similarity baseline with no pragmatics outperforms most of the language
models on this axis (Section~\ref{sec:literal}). Adding a pragmatic,
RSA-style layer over two similarity-based baselines moves them toward random
rather than toward the Bayesian reference (Section~\ref{sec:negative}), and
two of the four candidate baselines cannot supply that layer's required
literal semantics for structural reasons unrelated to coverage or noise.
\item We report a labelled multiple-choice response format, standard in
evaluation practice, that carries no measurable content signal in this setting:
permutation consistency is at or below chance in all 32 model-by-tile-by-label
cells (Section~\ref{sec:negative}).
\end{itemize}

Everything above is measured on the model side of a pre-registered instrument;
the fit function, the item set, the scoring rule, and the sensitivity battery
were all fixed before any model was run, and every model-side number is
reported as post hoc against that fixed instrument. The human side of the
instrument is built and piloted but not yet collected (Section~\ref{sec:discussion}).
The central claim, that item-sensitivity is not evidence of alignment without an
independent reference, is a statement about the model results already in hand
and does not depend on what the human data will show.

\section{Related Work}

\subsection{Generalization and similarity}
The fit function is the exponential form of Shepard's universal law of
generalization \cite{shepard1987}, applied to psychological scales, which is why
it carries no fitted decay rate. Nosofsky's GCM \cite{nosofsky1986} uses
Gaussian similarity for integral dimensions and exponential similarity for
separable ones, so the family is stimulus-dependent even within that literature.
We therefore treat the decay family as a reported dimension and run the entire
analysis under both forms rather than choosing one.

\subsection{Rational speech acts, and what is and is not ours}\label{sec:rsa-relwork}
Frank and Goodman \cite{frank2012} formalise a pragmatic speaker that trades
informativeness against distractor confusability. That trade-off is theirs, not
ours: the relationship between $o_{\text{fit}}$ and $o_{\text{bayes}}$ we use it
to define (Section~\ref{sec:oracle-diverge}) is an identity that follows from
the fit function's construction, not an empirical finding, and publishing it as
a discovery would be a serious error.

What is ours is narrower and, we think, more useful. We build a setting in which
the trade-off is exactly computable, because the hypothesis space is small
enough to enumerate and the likelihood is a closed-form function of
human-derived ratings, and we then measure which criterion each population
actually follows. The domain-specific quantities are the conflict rate, which
ranges from 3.78 percent to 37.82 percent by tile, and the magnitude, whose
median local competition ratio is 1.842.

\subsection{Concept norms and similarity data}
THINGS \cite{hebart2019things} supplies the concept universe, THINGSplus
\cite{stoinski2023thingsplus} supplies the property ratings from which tiles are
built, and SPoSE \cite{hebart2020spose}, fit to the odd-one-out triplets of
\cite{hebart2023data}, supplies a dense human-derived similarity space. The
distinction matters for our design. SPoSE is a compression of human judgements
rather than a language model, so using it to define the independent variable
does not make the analysis circular. The test is provenance, not format: a
66-dimensional vector is admissible if it was fit to human judgements and
inadmissible if it was fit to text. We describe it as a human-derived embedding
and never as ``an embedding''.

\subsection{Word association}
Condition 2 uses SWOW \cite{dedeyne2019swow}. Its cue list was built by snowball
sampling from association and feature-production norms, then expanded with
frequent responses, and that construction is the measured mechanism behind our
vocabulary's natural-kind retention bias rather than merely a plausible
explanation for it (Appendix~\ref{app:bias}). A concept enters the cue list
largely by being the kind of word people produce in free association, which
corpus frequency does not measure and SWOW's own response frequency does.

\subsection{Covariates and symbolic knowledge}
We carry concreteness \cite{brysbaert2014concreteness} and subtitle frequency
\cite{brysbaert2009subtlex} as covariates, using the columns already sense
disambiguated within THINGS rather than string-matching the standalone norms,
which would reintroduce the homograph ambiguity those columns resolve. Condition
3 uses ConceptNet \cite{speer2017conceptnet}.

\subsection{Human-model comparison and cooperative word games}
Hu et al. compare human and language-model performance on pragmatic language
understanding using a shared instrument administered to both populations
\cite{hu2023pragmatic}. Our design, in which the intended human condition
receives the identical stimulus a model receives and produces a response scored
on the same coordinate, is built in that tradition, applied to a single
signalling task with a closed-form reference rather than a battery of tasks
scored by human judgement.

Stephenson et al. use the party game Codenames as a benchmark for language
models \cite{stephenson2024codenames}, with clue-givers generating a free
one-word clue over an effectively open vocabulary. That openness is realistic
and is also why an exact Bayesian oracle or an exact random-selection reference
is not available there: the hypothesis space of possible clues cannot be
enumerated. Our menu trades that openness for a closed, fixed option set of 3
to 8 items specifically so that the salience, Bayesian, and random references
in Section~\ref{sec:align} are computable exactly rather than estimated.

The closest prior work to our design is Shaikh et al., who build the Cultural
Codes dataset from Codenames Duet, a cooperative two-player word game,
collecting 794 games (7,703 turns, 153 players) together with each player's
sociocultural background, and show that modelling a player's sociocultural
priors jointly with the game context improves prediction of both clue-giving
and guessing \cite{shaikh2023codenamesduet}. Their setting and ours share the
cooperative structure and a pragmatic-speaker lineage, but differ in what is
held closed and what is left open. Codenames Duet's clue vocabulary is
effectively unconstrained, so the space of possible messages cannot be
enumerated, and their account of pragmatic reasoning is fit to predict which
clue a player chose among many rather than compared against an exactly
computable reference. Our menu is a small, fixed, closed set chosen so that a
fit-maximising strategy, a posterior-maximising strategy, and uniform random
selection are all closed-form coordinates a model's choice can be measured
against exactly (Section~\ref{sec:definitions}). We also study a single
homogeneous literal-listener account rather than heterogeneity across human
sociocultural backgrounds, which is Shaikh et al.'s central question and is
outside the scope of the model-side results reported here.

\section{Methods}

\subsection*{What is pre-registered and what is not}
The instrument was fixed before any model was run. The fit function, the
aggregation over the target pair, the oracle, the item set and its selection
gates, the scoring rule, and the sensitivity battery are all pre-registered in a
frozen document. Every model-side measurement reported here is post hoc. The
measurements are post hoc; the instrument is not. Amendments made after model
data existed are listed in the frozen document's own disclosure and in a dated
amendment.

\subsection{Setting}\label{sec:setting}
The task is abstracted from the board game \emph{Deception: Murder in Hong
Kong}, designed by Tobey Ho and published by Grey Fox Games. In that game one
player knows a hidden pair of cards, a murder weapon and a piece of evidence,
and may communicate it to the other players only by placing markers on scene
tiles, each of which offers a small fixed set of categorical options that was
not written with those particular cards in mind. We take only that structure.
No card text, tile art or rulebook wording is reproduced here, and all
semantic content in our items comes from the open datasets described below.

The task is abstracted from a cooperative signalling problem embedded in an
adversarial audience. One party knows a hidden target pair of object concepts,
$(\textit{Means}, \textit{Clue})$, and may communicate it only by selecting one
option from a fixed menu that was not designed to express it. We call this
forced lossy encoding: choosing the least-bad symbol from a closed set.

An item presents $M$ candidate Means, $C$ candidate Clues, and a single
\emph{tile}, which is a categorical dimension with an option set $O$. The
hypothesis space is every $M \times C$ pairing, exactly one of which is the
target. Grid distractors rather than independently sampled ones are what create
the confusion an option must resolve. Uniform distractors are independent of the
target and generate no discrimination pressure. We use $M = C = 10$, giving 100
hypotheses. Each item carries one tile. This removes the conditional
independence assumption that multiplying likelihoods across correlated tiles
would require, and it makes the oracle assumption-free.

\subsection{Concept vocabulary}
Concepts come from THINGS \cite{hebart2019things}, with property norms from
THINGSplus \cite{stoinski2023thingsplus}. The vocabulary universe is the
intersection, computed before any filtering, of THINGS concepts that have
dimension ratings, triplet coverage, a Small World of Words cue match
\cite{dedeyne2019swow}, and non-null concreteness and frequency. No concept
enters that some condition cannot score, because a condition scored on a
different subset is not a comparison.

SWOW matching is exact-case string equality on the THINGS \texttt{Word} field,
with no lemmatisation, stemming or fuzzy matching, and with homographs excluded
on both sides. A cue shared by two concepts blends both senses, which is
semantic contamination rather than a coverage gap. Case-folding would admit nine
further concepts, two of which were collected as personal names, so the cost
asymmetry argues against it. The SWOW overlap is the binding constraint at
roughly 67 percent. Applying only a part-of-speech filter, which is definitional
rather than a tuning knob because THINGS pre-screened for concreteness,
familiarity and nameability, yields a pool of 1{,}118 noun concepts.

Our vocabulary carries a measured sampling bias toward natural kinds, whose
mechanism we trace to how the SWOW cue list was built. We report it in
Appendix~\ref{app:bias} because it describes our sample rather than our result.

\subsection{Tiles}
Tiles are property tiles rather than category tiles. THINGSplus typicality is
defined only where category membership exists, which leaves roughly 97 percent
of the concept-by-category matrix empty, and fit off the membership diagonal is
exactly the region the design must measure. Treating missing typicality as zero
would be false: a hammer is a decent weapon and would score zero. Property
ratings are defined for every concept on every dimension, so fit is continuous
everywhere.

We selected four dimensions by taking, within each redundant cluster
($|r| > 0.7$), the dimension with the lowest rater-disagreement ratio, defined
as mean within-concept SD divided by between-concept SD. File order was not a
criterion. The four are \texttt{manmade} (0.379), \texttt{size} (0.409),
\texttt{hold} (0.658) and \texttt{moves} (0.735), and the largest pairwise
correlation among them is $-0.63$. Three further dimensions were excluded on two
independent grounds, either of which is sufficient. \texttt{Arousal},
\texttt{precious} and \texttt{pleasant} have within-concept rater SD exceeding
between-concept SD, at ratios of 3.53, 1.62 and 1.14, so an option set built on
them would sort rater noise rather than concept differences. All three are also
evaluative rather than descriptive, and an option meaning ``the object was
pleasant'' is not the signalling task.

Bin boundaries come from the rating instrument and never from the pool. The
paper studies a menu that was not designed for the
messages it must carry, so placing boundaries where our concepts happen to
cluster would build a channel well matched to the message space, and the absence
of that match is the object of study. Likert dimensions bin on the nominal 1 to
7 range. \texttt{Size} bins on the instrument's own anchor objects at uniform
60-unit spacing, and the six size options are the anchor phrases verbatim. End
bins take the same width as interior bins, because width is the denominator of
the fit function and unequal widths would inflate fit at the extremes as a
measurement artifact. Bins that are empty or thin by construction are acceptable
and informative, since an option that fits nothing in the pool is a real
property of a fixed menu.

\subsection{Fit}
For an option $o$ with interval $I_o$, and a concept with value $v$,
\begin{equation}
f(o, x) =
\begin{cases}
1 & v \in I_o,\\[2pt]
\exp\!\left(-\,\mathrm{dist}(v, I_o) / \mathrm{width}(o)\right) & \text{otherwise,}
\end{cases}
\end{equation}
with the width taken from the instrument, so the function has no free parameter.
The exponential form is the theoretically motivated one: generalization decays
approximately exponentially with distance in psychological space
\cite{shepard1987}, and these are psychological scales. A linearly truncated
alternative censors straddle magnitude, because every pair spanning more than
one bin width scores exactly zero, so a pair 1.2 widths apart and one 4 widths
apart would record identically.

Fit for a pair is the minimum over its two members. Both \texttt{max} and
\texttt{centroid} return 1.0 for every target on every tile, which collapses the
design to a line, and \texttt{centroid} additionally selects an option that
describes neither member on 100 percent of straddling pairs. Taking the mean
discards straddle magnitude, which is the quantity the design exists to vary.
Under the minimum, fit is strictly positive everywhere, so the normalising sum
is never zero and no fallback distribution is ever introduced. A uniform
fallback is permanently rejected, because it would make the least describable
pairs indistinguishable from uniformly describable ones.

$P(\text{option} \mid \text{hypothesis})$ is the primitive, and target fit and
distractor attraction are derived from it. The two therefore cannot contradict
the oracle about which option is best.

\subsection{Oracle}\label{sec:oracle-diverge}
The oracle enumerates the exact posterior over all 100 hypotheses under a
uniform prior. Where two options are exactly tied, $o_{\text{bayes}}$ is defined
lexicographically as maximum posterior, then maximum fit among the options
achieving it. The tie is neither rare nor arbitrary, and its algebraic condition
is given in Appendix~\ref{app:ties}.

The oracle is an S1 speaker addressing a literal listener. It does not model a
listener who reasons about the speaker's strategy.

\paragraph{Why $o_{\text{fit}}$ and $o_{\text{bayes}}$ can differ}
This is load-bearing for the primary dependent variable defined in
Section~\ref{sec:definitions}, so we state it here rather than only in Related
Work. With a uniform prior,
\begin{equation}
P(h^* \mid o) = f(o,h^*) \,\big/\, {\textstyle\sum_h} f(o,h),
\end{equation}
$o_{\text{fit}}$ maximises the numerator and $o_{\text{bayes}}$ maximises the
ratio, so wherever the two differ, the denominator, which sums $f(o,h)$ over
every hypothesis and so measures how much that option also fits the
distractors, is necessarily smaller at $o_{\text{bayes}}$. Fit-maximising and
posterior-maximising options diverge exactly when an option's fit to
distractors, not only to the target, varies across the option set. This holds
by construction of the fit function and the uniform prior; it is an identity,
not an empirical finding, and Section~\ref{sec:rsa-relwork} states why
publishing it as a discovery would be a serious error.

\subsection{Items}
We generated 200{,}000 candidate items and assessed robustness in a separate
pass, so that the generator stays inspectable before filtering and a threshold
change does not force regeneration. Selection uses item properties only and
never model performance. The selection rule is declared, and the unselected
distribution is published alongside the selected one. The final set is 1{,}000
items, tile-balanced, half conflict and half control.

The measured item property is decision conflict, meaning that the option
maximising target fit differs from the option maximising the listener's
posterior. It is not position in a fit-by-attraction grid, which describes only
the chosen option and cannot detect conflict. Conflict rates are domain
specific, at 3.78 percent on \texttt{manmade} and 37.82 percent on
\texttt{size}. The median local competition ratio is 1.842.

\subsubsection{Definitions}\label{sec:definitions}
Write $h^*$ for the true target pair, $O$ for the item's option set, and
$f(o,h)$ for the pair fit of option $o$ under hypothesis $h$. Then
\begin{align}
o_{\text{fit}}   &= \arg\max_{o \in O} f(o, h^*),\\
o_{\text{bayes}} &= \arg\max_{o \in O} P(h^* \mid o),
\end{align}
with $P(h^*\mid o) = f(o,h^*)/\sum_{h} f(o,h)$ under the uniform prior, and
ties in $o_{\text{bayes}}$ broken lexicographically as described above.

\begin{figure*}[t]
\centering
\includegraphics[width=\textwidth]{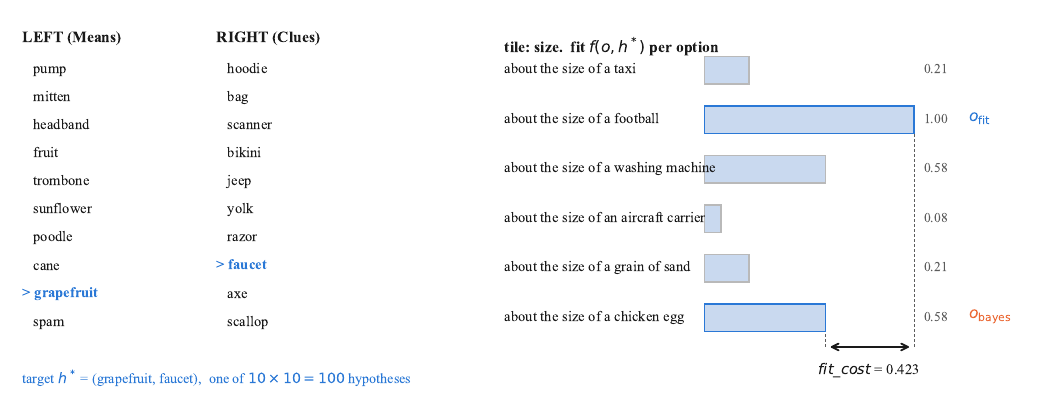}
\caption{One item. The chooser knows the target pair $h^*$ = (grapefruit,
faucet), marked on the two lists, and must send one of the six size options. The
bars are $f(o, h^*)$, the fit of each option to the target. $o_{\rm fit}$ is
``about the size of a football'', which describes the target best at 1.00.
$o_{\rm bayes}$ is ``about the size of a chicken egg'', which describes it worse
at 0.58 but discriminates it better from the other 99 hypotheses. The gap
between them is \textit{fit\_cost}, here 0.423: what a chooser gives up in
description quality by discriminating.}
\label{fig:example}
\end{figure*}

The primary dependent variable is the position of a chosen option $a$ on two
axes, each min-max normalised within that item's own option set:
\begin{align}
\textit{fit\_norm}(a)  &= \frac{f(a,h^*) - \min_{o} f(o,h^*)}
                               {\max_{o} f(o,h^*) - \min_{o} f(o,h^*)},\\
\textit{post\_norm}(a) &= \frac{P(h^*\mid a) - \min_{o} P(h^*\mid o)}
                               {\max_{o} P(h^*\mid o) - \min_{o} P(h^*\mid o)}.
\end{align}
By construction $\textit{fit\_norm}(o_{\text{fit}}) = 1$ and
$\textit{post\_norm}(o_{\text{bayes}}) = 1$, so a chooser that always selects
$o_{\text{fit}}$ sits at the salience pole and one that always selects
$o_{\text{bayes}}$ sits at the Bayes pole. Fig.~\ref{fig:example} shows one
item with both marked, and the two poles are easiest to build intuition for by
walking through what each chooser sends on it and why.

A fit-maximising chooser asks only how well an option describes the target,
regardless of what else it could describe. Of the six size options,
``about the size of a football'' fits grapefruit and faucet's true sizes most
closely, $f = 1.00$, so that is what such a chooser sends, and this is
$o_{\text{fit}}$. A posterior-maximising chooser asks a different question,
which of the options best distinguishes this particular pair from the other 99
candidate pairs the listener must also consider. ``About the size of a
football'' is a poor answer to that question: it is also true of 468 of the
1{,}118 pool concepts, the most crowded of the six size bins, so naming it
barely narrows down which pair was meant. ``About the size of a chicken egg''
fits the target a little worse, $f = 0.58$, but its bin holds 231 concepts,
roughly half as many, so sending it concentrates the listener's posterior over
hypotheses more than the better-fitting option does. That is what a
posterior-maximising chooser sends, and it is $o_{\text{bayes}}$.
\textit{fit\_cost} is exactly the fit given up, $1.00 - 0.58 = 0.42$, to buy
that concentration.

We define
$\textit{fit\_cost} = f(o_{\text{fit}}, h^*) - f(o_{\text{bayes}}, h^*)$
as the item difficulty variable and the x-axis of the divergence analysis. It is
what a speaker gives up in description quality by choosing to discriminate. The
posterior gap measures what the listener gains instead, and the two dissociate.
Where \textit{fit\_cost} is near zero, a salience-driven chooser and a
rationality-driven chooser are indistinguishable however large the gap.

\subsection{Rendering}
Concepts are presented as bare words. We removed glosses after a leak audit
found 206 leaking occurrences across 174 items, and because SWOW respondents saw
bare cue words, so glossing would introduce a presentation mismatch with the
baseline we compare against. The two candidate lists are disjoint by
construction, because a duplicated concept puts a self-pair into the hypothesis
space whose fit under the minimum is at least that of any genuine pair
containing it. No listed concept name may appear as a substring of any option
text on that item's tile. The rule is a pure substring test and deliberately
over-fires, because orthographic coincidence cannot be distinguished from
semantic priming without per-case adjudication, and a reader sees the letters
regardless of etymology. Option order is randomised per rendering, the
permutation is stored, and each item has two permutations.

\subsection{Scoring}
Models are scored by log-probability over the option set. There is no generation
path, so no generated string is ever parsed and no near-match tolerance exists.

Scores are PMI-normalised \cite{holtzman2021surface}, using one rule across
conditions 4 to 6:
\begin{equation}
\mathrm{score}(o) = \log P(o \mid \text{prompt}) - \log P(o \mid \text{neutral prompt}).
\end{equation}
A normalisation that varied across conditions would confound the comparison.
Normalisation is necessary rather than decorative. Within a single tile the
options differ in length by up to a factor of two, and on \texttt{moves} the
option ``moves'' is one token where ``stays still'' is three, so an unnormalised
sum favours the short option mechanically. We reject the per-token mean because
it trades a bias toward short options for a bias toward predictable ones.

The neutral prompt is the real menu with all item content stripped, in the order
that rendering displayed it. An earlier menu-free neutral prompt was wrong: a
model's prior over option strings exists only when a menu is present and varies
with the option count, so one table cannot serve 3-option, 4-option and
6-option tiles. Options are tokenized in place within the rendered prompt and
never standalone, because a leading space changes the token count.

We compute the unnormalised sum and the per-token mean in the same pass, and
they are not a footnote. The three rules agree on the chosen option only 58.0
percent and 63.4 percent of the time, which is the same magnitude of
methodological uncertainty as the choice of decay family. Every finding below is
therefore recomputed under all three rules and reported under all three, and any
finding that holds under one rule only is stated as rule-dependent in the
sentence that makes it. PMI is primary because it was fixed in advance, never
because it wins.

Ties are scored as the mean coordinate over the tied options, which is an
expectation under indifference rather than a tie-break, and array order plays no
part.

\subsubsection{The random reference}
Two poles bound the coordinate above but nothing bounds it below, so we add a
third reference: uniform selection over the option set. Its expected coordinate
is computed per item as the mean of $\textit{fit\_norm}$ and
$\textit{post\_norm}$ over that item's options, then averaged. On the 125
\texttt{size} conflict items this is
$\textit{fit\_norm} = 0.3985$ (SD 0.0294 across items) and
$\textit{post\_norm} = 0.4593$ (SD 0.0617). We report the spread because a
reference drawn as a single point invites the reader to treat it as exact.

\subsubsection{Intervals}
Intervals are 95 percent percentile bootstrap, 4{,}000 resamples, with the
\emph{item} as the resampling unit. Two permutations of the same item are not
independent observations, so the 125 items rather than the 250 renderings are
the clusters. Where we compare a model against the random reference we resample
the paired per-item difference. We report per-cell intervals and, where a count
across the 21 model by rule cells is at issue, we also state the count under a
Bonferroni correction.

\subsection{Response format}
We pre-registered two prompt formats, because forced-choice answers are known to
flip on trivial relabelling \cite{sclar2024format}. Format L, in which options are labelled and the
answer is a letter, is retired as a measurement condition and reported as a
result in Section~\ref{sec:negative}. All analyses use format V, in which the
answer repeats the option text. Appendix~\ref{app:ui} shows the interface built
for instrument testing.
\pending{the human response channel: a forced best guess followed by a separate
refusal question, with the interface's stated departure from stimulus identity.}

\subsection{Models and environment}
The ladder is Qwen3 at 0.6B, 1.7B, 4B and 8B \cite{qwen3}, chosen because scale
varies while the training recipe does not, which is what a scale claim requires.
The 1.7B and 8B base models give a post-training ablation, and OLMo-2-7B-Instruct
\cite{olmo2} is a cross-family control, preferred because its pretraining data
is open.

All runs are fp16 on two T4 GPUs, with pinned model revision hashes, tokenizer,
seed and library versions. Every rung runs at the same precision, because a
scale trend measured across mixed precisions cannot be distinguished from a
quantization artifact.

We measure one precision effect rather than removing it. Different batch sizes
select different GEMM reduction orders, which gives an fp16 noise floor of about
0.24 in summed log-probability on realistically padded batches. A diagnostic
separates precision from correctness: the equivalent fp32 check collapses to
$1.9 \times 10^{-6}$, which proves the padding and positional encoding are
correct, since a bug is not a precision effect and would survive fp32 at full
magnitude. fp32 is unavailable rather than declined, because Qwen3-8B in fp32 is
32~GB against roughly 30~GB usable, T4 is Turing so bf16 is not native, and
dropping one rung to another precision is forbidden. The correctness gate
therefore runs once in fp32, and the per-rung fp16 check is dropped rather than
loosened, because a check that cannot separate a bug from noise has no power.
The \texttt{size} tile, which carries the primary analysis, is scored at batch
size 1, so it has no padding and no batch effect. The remaining tiles are
batched, with a batch-1 subsample re-scored and the flip rate reported.

Two constraints govern cross-model comparison. Base models are scored with their
instruct sibling's chat template applied verbatim. Both alternatives confound
something, but the template's confound is one-directional and therefore
informative: if base models show less item-sensitivity that is not attributable,
whereas if they show the same or more, format unfamiliarity cannot explain it.
Separately, cross-family comparisons are restricted to choice-based statistics,
namely excess over the marginal null, permutation consistency, agreement rates
and position bias. Raw log-probability magnitudes are never compared across
families, because the ladder's comparability rested on a shared tokenizer that
the control breaks by construction. The run asserts that the tokenizer differs
rather than assuming it.

\subsection{Conditions}
The eight conditions are: (1) embedding cosine with no pragmatics, a mandatory
literal baseline that is never omitted; (2) SWOW random-walk association
\cite{dedeyne2019swow}; (3) ConceptNet path scoring \cite{speer2017conceptnet}
from a local dump; (4) LLM zero-shot; (5) as (4), with the hypothesis space made
salient by one added sentence; (6) an RSA layer; (7) the exact Bayesian oracle;
and (8) humans.
\pending{condition 8 is named here and not collected. It becomes a measured
condition, and Results gains the human coordinate on the same axes.}

Conditions 1 to 3 hold pair aggregation and word-level aggregation constant with
the rest of the design, so that comparing them isolates the similarity source
rather than confounding source with aggregation.

\section{Results}
Unless stated otherwise, results are on the 125 \texttt{size} conflict items,
which are the primary analysis set. Every figure is reported under all three
scoring rules, and means are reported with medians.

\subsection{Models respond to items}\label{sec:sensitive}
We measure item-sensitivity as \emph{excess}: the probability that a model makes
the same canonical choice for an item under both stored permutations, minus what
its own marginal option preference would produce. A uniform null would be wrong
here, because a model choosing one option 93 percent of the time scores 0.87
consistency while being entirely item-insensitive. This corrects an earlier
version of our own diagnostic.

On \texttt{size}, excess is positive in every one of the 21 model by rule cells,
covering seven models under three scoring rules with no exception. On the 125
conflict items it ranges from $+0.024$ to $+0.260$. It is the only tile with
this property, which we return to in Section~\ref{sec:negative}.

Read on its own, this says the models are doing the task. Every model responds
to the item rather than to its own option preference, under every scoring rule,
including the base models and the cross-family control. Section~\ref{sec:align}
shows what that does and does not buy.

\begin{figure*}[t]
\centering
\includegraphics[width=\textwidth]{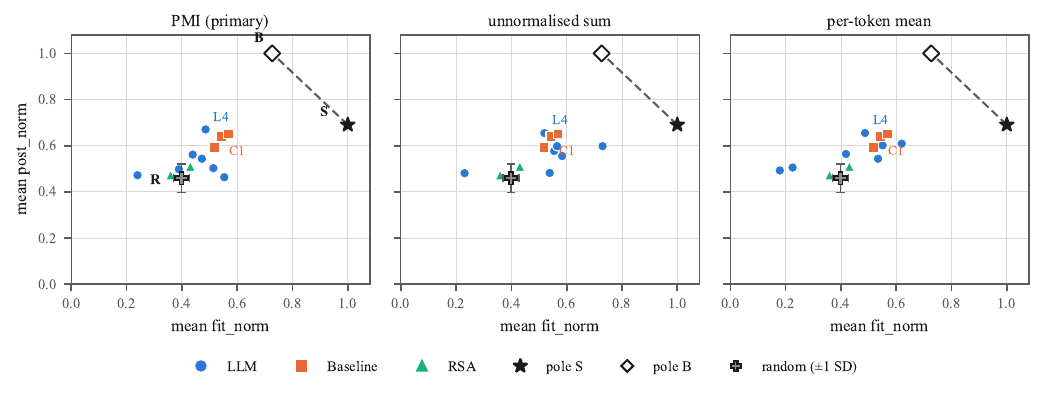}
\caption{Every chooser against three references, faceted by scoring rule. S is
the salience pole, B is the Bayes pole, and R is uniform random selection over
the option set, drawn with plus or minus one standard deviation across items.
Models occupy the band between random and salience. Baselines and RSA layers
have no scoring-rule dimension, so they repeat across panels. Colour marks the
family and marker shape repeats it, so identity is never carried by colour
alone.}
\label{fig:frontier}
\end{figure*}

\begin{table*}[t]
\centering
\caption{Coordinate summary per model and scoring rule, on the 125 \texttt{size} conflict items. Mean and median are raw $\textit{post\_norm}$ (random reference 0.459, salience pole 0.691, Bayes pole 1.000). Excess is item-sensitivity over the marginal null (Section~\ref{sec:sensitive}). $\Delta$random is the paired difference from the per-item random reference (Section~\ref{sec:align}); a dagger marks a 95 percent interval that excludes zero.}
\label{tab:results}
\resizebox{\textwidth}{!}{%
\begin{tabular}{lccccccccc}
\toprule
 & \multicolumn{3}{c}{PMI (primary)} & \multicolumn{3}{c}{unnormalised sum} & \multicolumn{3}{c}{per-token mean} \\
\cmidrule(lr){2-4} \cmidrule(lr){5-7} \cmidrule(lr){8-10}
Model & mean & median & excess & mean & median & excess & mean & median & excess \\
\midrule
Qwen3-0.6B & 0.498 & 0.500 & +0.048 & 0.555$^\dagger$ & 0.560 & +0.130 & 0.543$^\dagger$ & 0.558 & +0.109 \\
Qwen3-1.7B & 0.543$^\dagger$ & 0.507 & +0.080 & 0.577$^\dagger$ & 0.537 & +0.059 & 0.564$^\dagger$ & 0.530 & +0.094 \\
Qwen3-4B & 0.561$^\dagger$ & 0.525 & +0.090 & 0.598$^\dagger$ & 0.617 & +0.223 & 0.609$^\dagger$ & 0.579 & +0.182 \\
Qwen3-8B & 0.670$^\dagger$ & 0.740 & +0.260 & 0.654$^\dagger$ & 0.811 & +0.143 & 0.655$^\dagger$ & 0.862 & +0.221 \\
Qwen3-1.7B-Base & 0.463 & 0.482 & +0.097 & 0.482 & 0.506 & +0.133 & 0.505 & 0.500 & +0.024 \\
Qwen3-8B-Base & 0.503 & 0.500 & +0.154 & 0.481 & 0.500 & +0.216 & 0.493 & 0.499 & +0.182 \\
OLMo-2-7B & 0.472 & 0.500 & +0.146 & 0.598$^\dagger$ & 0.710 & +0.048 & 0.601$^\dagger$ & 0.710 & +0.069 \\
\midrule
random & \multicolumn{9}{c}{0.459 (SD 0.062 across items), no rule dimension} \\
\bottomrule
\end{tabular}}
\end{table*}

\subsection{Models align with neither criterion}\label{sec:align}
Each item defines two poles. The coordinate at $o_{\text{fit}}$ is where a
salience-driven chooser lands, and the coordinate at $o_{\text{bayes}}$ is where
a rationality-driven chooser lands. On this set the salience pole sits at
$\textit{post\_norm} = 0.691$ and the Bayes pole at 1.000. Uniform random
selection sits at 0.459, with a standard deviation of 0.062 across items.

We report raw $\textit{post\_norm}$ as the primary quantity and the
salience-to-Bayes fraction only as a secondary. The interval is 0.309 wide, so
per-item fractions are unstable where an item's own span is small, and an
earlier version of this analysis produced values of $-1$ to $-4$ on a quantity
bounded near $[0,1]$ before we moved off the ratio.

\paragraph{Models occupy the band between random and salience}
All 21 model by rule cells fall between the random reference and the salience
pole, from 0.463 to 0.670 (Fig.~\ref{fig:frontier}). None reaches the salience
pole and none approaches the Bayes pole. Expressed as a fraction of the gap from
random to salience, the cells run from $+0.02$ to $+0.91$, and the highest is
the 8B instruct model under PMI.

The earlier framing of this result, that models sit outside the
salience-to-Bayes interval, is true and nearly vacuous. Random selection sits at
$-0.75$ on that scale, further outside the interval than 20 of the 21 cells, so
a coin flip satisfies the claim. What the random reference buys is the ability
to say where in the band each model sits.

\paragraph{Eight of 21 cells are not distinguishable from random}
Comparing each cell against the per-item random reference, with the item as the
resampling unit, eight cells have a 95 percent interval that includes zero: all
six base-model cells, the cross-family control under PMI, and the 0.6B instruct
model under PMI. The paired differences there run from $+0.004$ to $+0.046$.
Under a Bonferroni correction across the 21 cells, 11 are not distinguishable.

This is a precision statement rather than an equality claim. At 125 items the
intervals are roughly $\pm 0.06$ wide, so we can say these cells are not
distinguishable from random and we cannot say they are equal to it.

\paragraph{Five of 21 cells are worse than random at describing the target}
On the fit axis the random reference is 0.399. Five cells fall below it: the
cross-family control under PMI at 0.239, the 8B base model under the
unnormalised sum at 0.231 and under the per-token mean at 0.179, the 1.7B base
model under the per-token mean at 0.225, and the 0.6B instruct model under PMI
at 0.389. This is a different claim from misalignment. These choosers select
options that describe the target worse than a coin flip would.

\paragraph{Mean and median say different things, and both are true}
Table~\ref{tab:results} reports both, on the item-clustered basis described
above ($n=125$). Mean and median diverge in both directions and by no
consistent amount: mean exceeds median in 7 of the 21 cells and median exceeds
mean in the other 14, so neither statistic is redundant with the other and
neither summarises a simple two-point distribution. Under item-clustering, no
model by rule cell has even half its items agreeing with $o_{\text{fit}}$ on
both stored permutations, which is the criterion for landing exactly at the
salience pole, so we do not describe the coordinate as bimodal at the poles. A
per-rendering count that treats the two permutations as independent gives a
different, larger figure, and we do not use it, for the same reason we resample
by item rather than by rendering in Section~\ref{sec:align}.

The 8B instruct model is the case where the direction matters. Its median is
0.74 to 0.86 across the three rules, above its own mean in all three and well
above every other model's median, so on the typical item it sits closer to the
Bayes pole than the other six models do on theirs. That is one rung differing
from three, and it is not a scale claim.

\subsection{Consistency is not alignment}\label{sec:thesis}
Sections~\ref{sec:sensitive} and \ref{sec:align} are in tension, and resolving
that tension is this paper's main claim.

Excess is positive in 21 of 21 cells. Eight of those same cells sit at a
coordinate that is not distinguishable from random. Every one of the eight has
positive excess, from $+0.024$ to $+0.216$, and the sharpest case is the 8B base
model under the unnormalised sum: its excess of $+0.216$ is the third highest of
all 21 cells, while its coordinate is $+0.022$ from random with an interval
spanning zero.

Across cells, excess and distance from random correlate at only $r = +0.30$.
Knowing that a model responds reliably to the item tells you very little about
whether it responds to anything the experiment is about.

We take this to generalise past this study. Excess, permutation consistency, and
the wider family of item-sensitivity and self-consistency metrics all measure
whether a model's output depends on the input rather than on its own output
prior. They are necessary conditions for a model doing a task, and they are
often reported as though they were sufficient. They are not. A model can be
strongly item-sensitive while tracking a feature of the item that the
experimenter does not care about, and no amount of consistency will reveal that.
Detecting it requires a reference for the quantity of interest, which is what
the random baseline supplies here and what the marginal null supplies for excess.

\begin{figure*}[t]
\centering
\includegraphics[width=\textwidth]{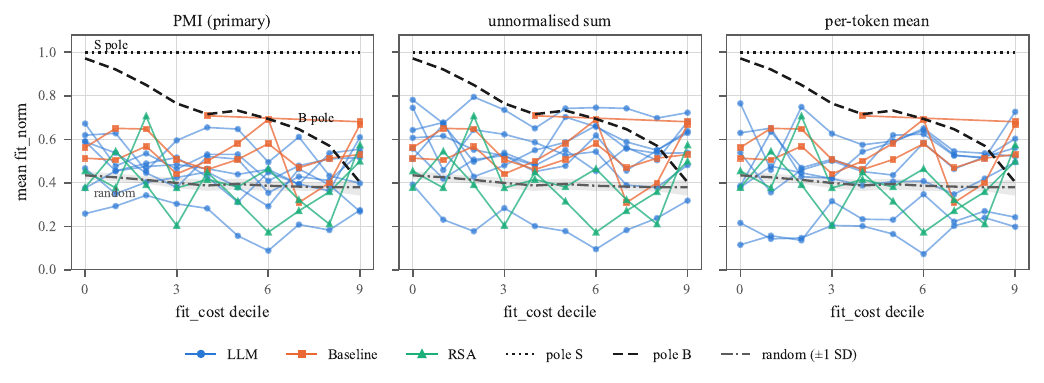}
\caption{Mean achieved fit against \textit{fit\_cost} decile, faceted by scoring
rule. The dotted line is what a salience chooser would score, the dashed line
what a Bayesian chooser would score, and the dash-dot band is uniform random
selection with plus or minus one standard deviation across items. Reference
slopes are 0.00 for salience and about $-0.90$ for Bayes. Observed model slopes
run from $-0.43$ to $+0.25$, so they are shallow rather than flat, and no
interval excludes zero under the primary rule.}
\label{fig:divergence}
\end{figure*}

\subsection{Achieved fit is the discriminating axis}
Because
$\textit{fit\_cost} = f(o_{\text{fit}}, h^*) - f(o_{\text{bayes}}, h^*)$,
the two strategies separate on it by construction, on the fit axis. A salience
chooser holds normalised fit at 1.000 regardless of \textit{fit\_cost}, whereas
a Bayesian chooser's achieved fit falls as \textit{fit\_cost} rises. Measured
across deciles, the Bayes pole falls from 0.972 to 0.403 and the separation
grows monotonically from 0.028 to 0.597. On the posterior axis the separation is
smaller and non-monotone, from 0.128 to 0.415, because the salience pole already
averages 0.691 there.

On this axis models sit between $\textit{fit\_norm}$ 0.09 and 0.67, against the
salience pole's 1.000 and the random reference's 0.399
(Fig.~\ref{fig:divergence}). The reference slopes are 0.00 for salience and about
$-0.90$ for Bayes, and observed model slopes run from $-0.43$ to $+0.25$. We
describe these as shallow rather than flat, because a slope of $-0.43$ is
roughly halfway to the Bayesian reference and calling it flat would overstate
the result. Only three of the 21 intervals exclude zero, all from one model and
with the sign reversing across rules, so we claim no slope for any model.

\subsection{The divergence curve is shallow under the primary rule}
Under PMI, no model's slope of mean coordinate on \textit{fit\_cost} has an
interval excluding zero. The 8B instruct model's predicted change across the
observed range is $+0.002$, against a pre-registered smallest effect of interest
of 0.05. Under the other two rules, five cells do have intervals excluding zero,
all positive. That is a rule-dependent result and is stated here as such.

This leaves an open question that our data can pose but not answer. Models are
item-sensitive on \texttt{size}, so their choices depend on which item they see,
yet their coordinate does not move with \textit{fit\_cost}, which is the item
property the design varies. Whatever they respond to is a feature of the item
uncorrelated with the fit cost of discriminating. We can rule out two
candidates. It is not the model's own option preference, which the marginal null
removes, and it is not position, because the coordinate is computed on the
canonical option and permutation consistency is above the marginal null.
Identifying what it is would require a manipulation the current item set does
not contain, so we state the question and leave it open rather than speculate.

We also report the rate at which a model departs from $o_{\text{fit}}$. That
rate is at a ceiling for most models, and the reason is structural rather than
incidental. At the lowest decile the two options differ by only 0.028 of fit, so
asking whether the chooser picked exactly $o_{\text{fit}}$ has no floor there.
The measure starts high because it is uninformative where \textit{fit\_cost} is
small. We report it, but it is not a basis for design decisions.

\subsection{The literal baseline outperforms most models}\label{sec:literal}
Condition 1, embedding cosine with no pragmatics, sits at
$\textit{post\_norm} = 0.640$, which is 78 percent of the way from random to the
salience pole. That is above 15 of the 21 model cells. Condition 3 reaches 83
percent and condition 2 reaches 58 percent. Our pre-registration requires
reporting condition 1 even if it wins, and on this measure it beats most of the
ladder. It is also the one base condition that cannot support a pragmatic layer,
as Section~\ref{sec:negative} shows. We report both facts together, because the
first alone reads as a recommendation and the second alone reads as a defect.

\subsection{Making the hypothesis space salient}
Condition 5 adds one sentence to condition 4, telling the chooser that the other
person will consider every possible pairing rather than only the one it has in
mind. It changes the chosen option on roughly one item in six, with agreement
between the two conditions at 0.834 overall.

The effect is concentrated in one model. Agreement is 0.888, 0.934 and 0.893 for
the 0.6B, 1.7B and 4B instruct models, and 0.621 for the 8B. On \texttt{manmade}
the 8B model's agreement falls to 0.288 and the total variation distance between
the two conditions' option distributions reaches 0.690, so the added sentence
relocates most of the probability mass. The same caution applies as elsewhere.
This is one rung differing from three under one grouping, and we do not read it
as scale.

\subsection{Negative results}\label{sec:negative}
These are results, not limitations. Limitations are stated separately in the
pre-registration. A Results section organised around what worked would bury
them, and we take them to be among the paper's stronger contributions.

\subsubsection{Two conditions are structurally ineligible for a pragmatic layer}
RSA requires a literal listener, $L_0(h \mid o)$. A prompted language model
supplies $P(o \mid \text{prompt})$ with the prompt naming the target, which is a
speaker likelihood, and deriving a listener from it by Bayes assumes the model
is a rational speaker, which is what the layer exists to test. Separately,
cosine similarity is signed, unbounded below, and carries no probabilistic
interpretation: 12.1 percent of hypothesis entries are negative and
$L_0^{\alpha}$ is undefined at $\alpha = 0.5$. We rejected every transform.
Clipping at zero is parameter-free but not assumption-free, because it asserts
that a negative cosine means zero compatibility elementwise, where our existing
convention applies at the maximum. Shifting and softmax both reintroduce a free
parameter. Of the four base conditions we nominated, two are ineligible, for two
different structural reasons, and neither for a reason that coverage or noise
would suggest. What a pragmatic layer can be built on is narrower than the
literature's usage implies.

\subsubsection{The pragmatic layer moves choosers toward random}
With the random reference in place this result reads more sharply than we first
stated it. RSA over SWOW moves $\textit{post\_norm}$ from 0.593 to 0.510, and
RSA over ConceptNet from 0.651 to 0.472. On the scale from random to salience
that is a fall from 58 percent to 22 percent, and from 83 percent to 6 percent.
\textbf{RSA over ConceptNet ends up indistinguishable from a coin flip.}

The question was not whether the layer moves choosers from salience toward
Bayes, but whether it moves them into the interval at all. It moves them the
other way, and specifically toward random rather than merely away from the
poles. Coverage differs between a base condition and its layer, since RSA over
ConceptNet scores 116 items where ConceptNet itself scores 55, and we never
table the two together without stating that.

\subsubsection{Three of four tiles carry no item-sensitivity}
Only \texttt{size} shows excess over the marginal null. Under PMI,
\texttt{moves} is below its own marginal at all four ladder rungs, meaning the
choice moves with the permutation more than independent sampling would produce,
and \texttt{hold} carries approximately nothing. This removes the low-end
replication the design planned, so the primary curve rests on \texttt{size}
alone. The specific mechanism is rule-dependent, since under the unnormalised
sum and the per-token mean neither the \texttt{moves} nor the \texttt{hold}
statement holds, while the conclusion that \texttt{size} is the only carrying
tile survives all three rules.

\subsubsection{The labelled response format carries no content signal}
We pre-registered format L as the control on the normalisation, predicting that
after correct normalisation the two formats should largely agree. It falsified.
Permutation consistency is at or below chance in all 32 model by tile by
label-form cells, so the canonical choice is statistically independent of the
item. Eleven cells select a single letter on 100.00 percent of renderings, and
every canonical distribution sits within 0.10 of uniform. The two label forms,
\texttt{(A)} and bare \texttt{A}, agree on only 64.9 percent of choices. Format
L measures the model's letter prior and nothing else. This is a replication of a
known effect, that multiple-choice language model evaluation is not robust to
answer position and labelling \cite{zheng2024mcq}, in a setting where we can
additionally show the canonical choice carries no information at all rather
than only a positional bias. We report it as the outcome of a pre-registered
prediction that failed, which is what pre-registration is for.

\subsubsection{Two parameters cannot affect an argmax}
RSA's rationality parameter is inert. Since $x^{\alpha}$ is strictly monotone
increasing on $x \ge 0$ for $\alpha > 0$, the exponent cannot reorder a
non-negative vector, and the chosen option is identical across the swept range
in 100.0000 percent of 2{,}000 base by item cells. Independently, condition 3's
decay base is inert for the same reason at a different point in the pipeline,
because $\arg\max_o d^{K(o)} = \arg\min_o K(o)$ for all $d \in (0,1)$. These are
one finding stated twice. A parameter that looks like a reported dimension and
is algebraically incapable of affecting the measure is a recurring property of
this design rather than two coincidences. We never present four identical rows
as convergent evidence. Both parameters remain live for any distributional
report, where they sharpen rather than reorder.

\pending{Results gains the human coordinate on the same axes, the human
divergence curve, which is the confirmatory test the model arm no longer is, and
an individual-differences analysis if the two-strategy reading from instrument
testing survives at scale.}

\section{Discussion}\label{sec:discussion}
The central claim, that item-sensitivity is necessary and not sufficient for
task alignment, is supported by the model-side results alone and is not
contingent on what follows. The limits below bound what else can be concluded
from this paper as it stands.

\paragraph{No human data yet}
\pending{this paragraph is written for the model-only paper and is superseded
once condition 8 is collected.} Condition 8 is designed and piloted for
instrument testing only, and not collected. The random and salience references
in this paper describe two idealised strategies; we do not yet know where a
human chooser falls between them, and the paper's design question, does either
population track either criterion, is answered here only for models. The
pre-registered instrument includes both a block powered to classify an
individual participant's strategy and a block powered to estimate a population
curve, because a population mean can describe nobody if strategies vary across
people; we treat that as a possibility the design should be able to detect,
not as an established finding.

\paragraph{One tile carries the primary analysis}
Of the four tiles in the design, only \texttt{size} shows item-sensitivity
above the marginal null (Section~\ref{sec:negative}); the other three do not
replicate it, for reasons that are themselves rule-dependent. The
\textit{fit\_cost} analysis, the frontier position results, and the results
table are consequently reported on \texttt{size} alone. We do not know whether
the consistency-without-alignment result is a property of this task in general
or is concentrated on the one tile with enough resolution to detect either
strategy at all; the coarser tiles may show the same dissociation without our
being able to measure it.

\paragraph{The oracle is a literal-listener account}
$o_{\text{bayes}}$ is the optimum for an S1 speaker addressing a listener who
does not reason about the speaker's strategy. A model, or a person, reasoning
about how the listener will interpret their own choice would not necessarily
converge on $o_{\text{bayes}}$, and our claim that models sit far from the
Bayesian reference should be read as a claim about this literal-listener
account specifically, not about pragmatic rationality in general. Condition 5
tests whether making the hypothesis space explicit moves models, and it does
move one of them substantially (Section~\ref{sec:negative}); a fuller
higher-order account is outside this paper's scope.

\paragraph{The vocabulary carries a measured sampling bias}
Appendix~\ref{app:bias} reports a retention gap toward natural kinds and traces
it to how the association norms underlying one of our conditions were
constructed. We report it because it describes our sample, not because we
believe it drives the results in Section~\ref{sec:align}, which are model
properties rather than vocabulary properties; but a vocabulary drawn differently
could in principle populate the item set differently, and we have not run that
check.

\paragraph{What would change this paper's central claim, and what would not}
A different scoring rule showing all 21 cells clearly aligned with either
reference would change it; we tested three rules and found none does. A
different tile showing item-sensitivity without the random-indistinguishable
cells would narrow it to the tiles we happened to pick with too little
resolution, not remove it. Human data showing people also land indistinguishably
from random on the harder items would extend the claim to a second population;
human data showing people cleanly separate the two strategies would sharpen the
contrast with models without weakening the model-side finding, which does not
reference human behaviour to begin with. We do not see a result on the model
side, obtained under the pre-registered instrument, that this paper's central
claim depends on and that we consider likely to reverse.

\section{Conclusion}
Across seven language models, two model families, a post-training ablation, and
three scoring rules, every model is item-sensitive on the tile that carries our
primary analysis: every one of 21 model-by-rule cells responds to which item it
is shown. Yet eight of those same 21 coordinates are not statistically
distinguishable from a chooser that ignores the item and selects at random, and
five describe the target worse than random selection would. Item-sensitivity
and distance from a random reference correlate at only $r=+0.30$. We call this
consistency without alignment: a widely used class of evidence that a model's
output depends on the input is necessary but not sufficient to show the model is
doing the task the experimenter intends, and demonstrating the gap requires an
independent reference for the quantity of interest rather than only a null for
the metric itself.

The human arm of this instrument is designed and pre-registered but not yet
collected. It will place a second population's choices on the same
fit-posterior-random coordinate system used throughout this paper, test whether
people separate into fit-maximising and posterior-maximising strategies where
models do not, and extend the \textit{fit\_cost} interaction, pre-registered as
confirmatory for humans and reported here only as descriptive for models, to a
population where it has not yet been measured.

\section*{Acknowledgment}
This paper's setting originated in conversations, while playing \emph{Deception:
Murder in Hong Kong} with friends, about whether a fixed menu of descriptions
can carry an arbitrary message and how hard it is to build a chain of
associations under that constraint. \emph{Deception: Murder in Hong Kong} was
designed by Tobey Ho and published by Grey Fox Games. We thank the friends who
played it with us for that conversation, without naming them here.

\section*{Data Availability}
Released in the accompanying repository: the final item set (1{,}000 items,
both stored permutations), the rendered prompts, the scored model choices under
all three scoring rules, the full decision log, and all code used to generate,
render, score, and analyse them. Not released: a redistributed copy of any
source corpus. The Small World of Words association norms
\cite{dedeyne2019swow} are licensed CC BY-NC-ND, which blocks redistribution
outright; we release only the scores and choices computed from them. ConceptNet
\cite{speer2017conceptnet} is licensed CC BY-SA 4.0, which permits
redistribution under attribution and share-alike; we do not mirror our local
dump and instead point to its public source. THINGS \cite{hebart2019things} and
THINGSplus \cite{stoinski2023thingsplus} are released by their authors under a
CC0 public-domain dedication, which places no restriction on redistribution; we
do not mirror the raw archive here and instead point to its canonical OSF
repository (project \texttt{jum2f}). Every corpus we use can be reconstructed
by a reader from its own public source and the acquisition steps in our data
manifest.

\section*{Ethics Statement}
No human participants were recruited, and no participant data is reported in
this paper; every result reported here is on the model side of a
pre-registered instrument. Two colleagues reviewed the instrument for clarity
and wording during its development; this review was instrument development
rather than human-subjects research, none of their responses is reported
here, and they are not named. The human condition is designed and
pre-registered but not yet collected, and its collection will proceed under
informed consent through the Prolific platform.

\appendices
\section{A measured sampling bias in the vocabulary}\label{app:bias}
Natural kinds are retained at 87.1 percent against artifacts at 62.1 percent, a
gap of 25.0 percentage points. Conditioning on single-word concepts leaves 18.5
points, so multi-word lexicalisation explains only 26 percent of it, and within
the rarest corpus-frequency quartile the gap is 26.4 points, which is as large
as the unconditioned gap.

The mechanism is associative availability. SWOW's cue list was built by snowball
sampling seeded from association and feature-production norms, then expanded
with words that appeared frequently as responses. Every one of the 1{,}245
cue-list concepts has been produced as a response at least once, while 21.0
percent of non-cue-list concepts never have. Response frequency subsumes corpus
frequency: in a model of cue-list membership, corpus frequency falls from
$+2.95$ ($p<0.001$) to $+0.47$ ($p=0.066$) once response frequency is included,
and pseudo-$R^2$ rises from 0.469 to 0.722. The artifact term is reduced by 35
percent but not eliminated.

This describes our vocabulary's sampling and is never presented as a test of the
research question.

\section{The exact tie condition}\label{app:ties}
Two options tie exactly when their likelihood columns over the hypothesis space
are proportional. The geometric statement, that both options lie entirely
outside the item's concept range, is sufficient but not necessary and covers
84.7 percent of ties. The remainder arises when the member that binds the
minimum lies outside both options while the other does not. We test the
algebraic condition, of which the geometric one is a special case.

The reason ties occur at all is that the oracle is exactly indifferent among
options lying wholly outside the range of an item's concepts. For any such
option, every concept's distance differs from its distance to the next option
out by a constant $c$. Since $f = \exp(-\mathrm{dist}/\mathrm{width})$ with
uniform widths, every fit is scaled by $\exp(-c/\mathrm{width})$, the minimum
aggregation preserves the common factor, and the posterior normalises it away.
This holds exactly rather than approximately.

\section{The annotation interface}\label{app:ui}
\begin{figure}[t]
\centering
\includegraphics[width=\linewidth]{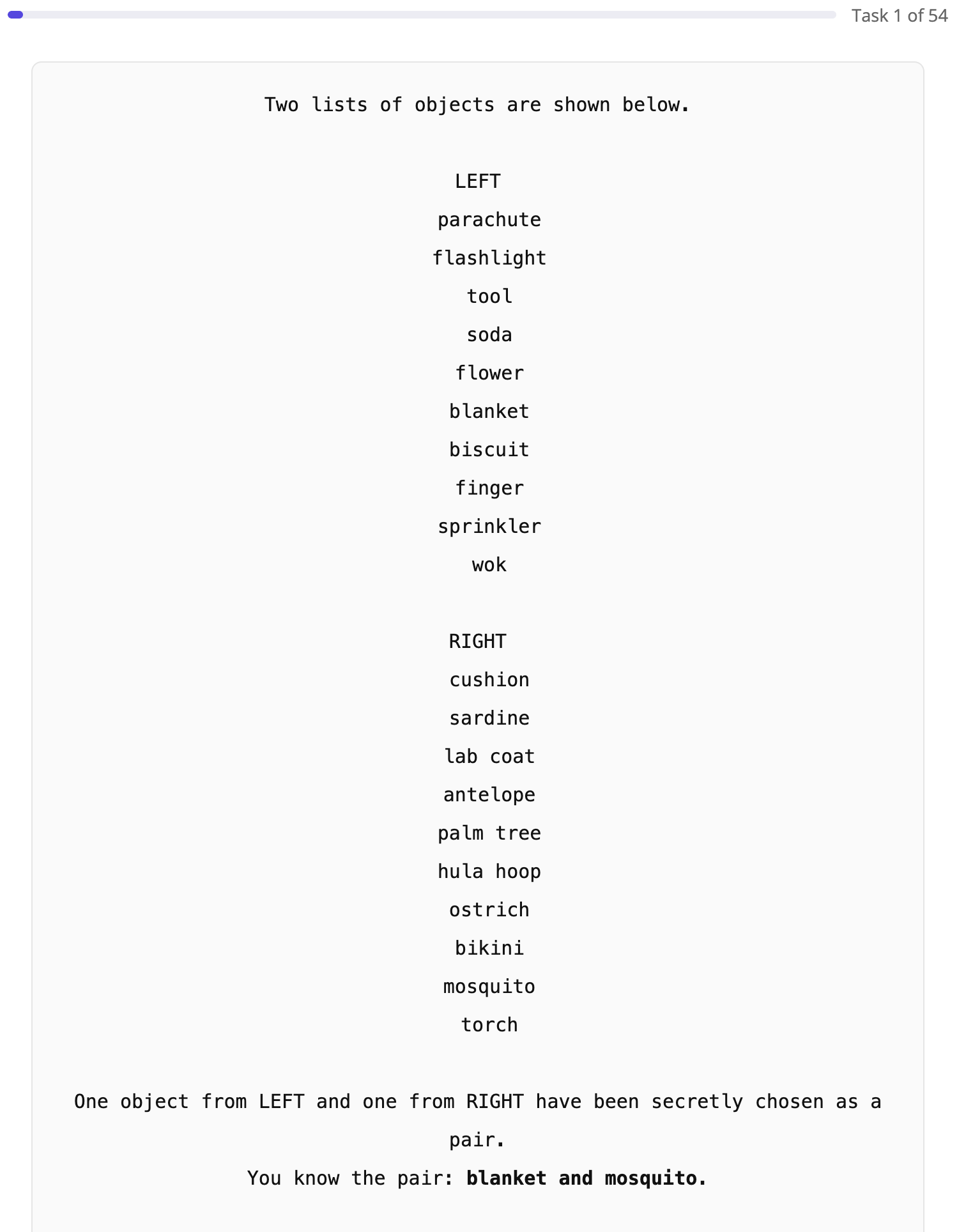}
\caption{The item as shown to a chooser: the two candidate lists with the
target pair marked below them. No participant data is shown.}
\label{fig:ui1}
\end{figure}

\begin{figure}[t]
\centering
\includegraphics[width=\linewidth]{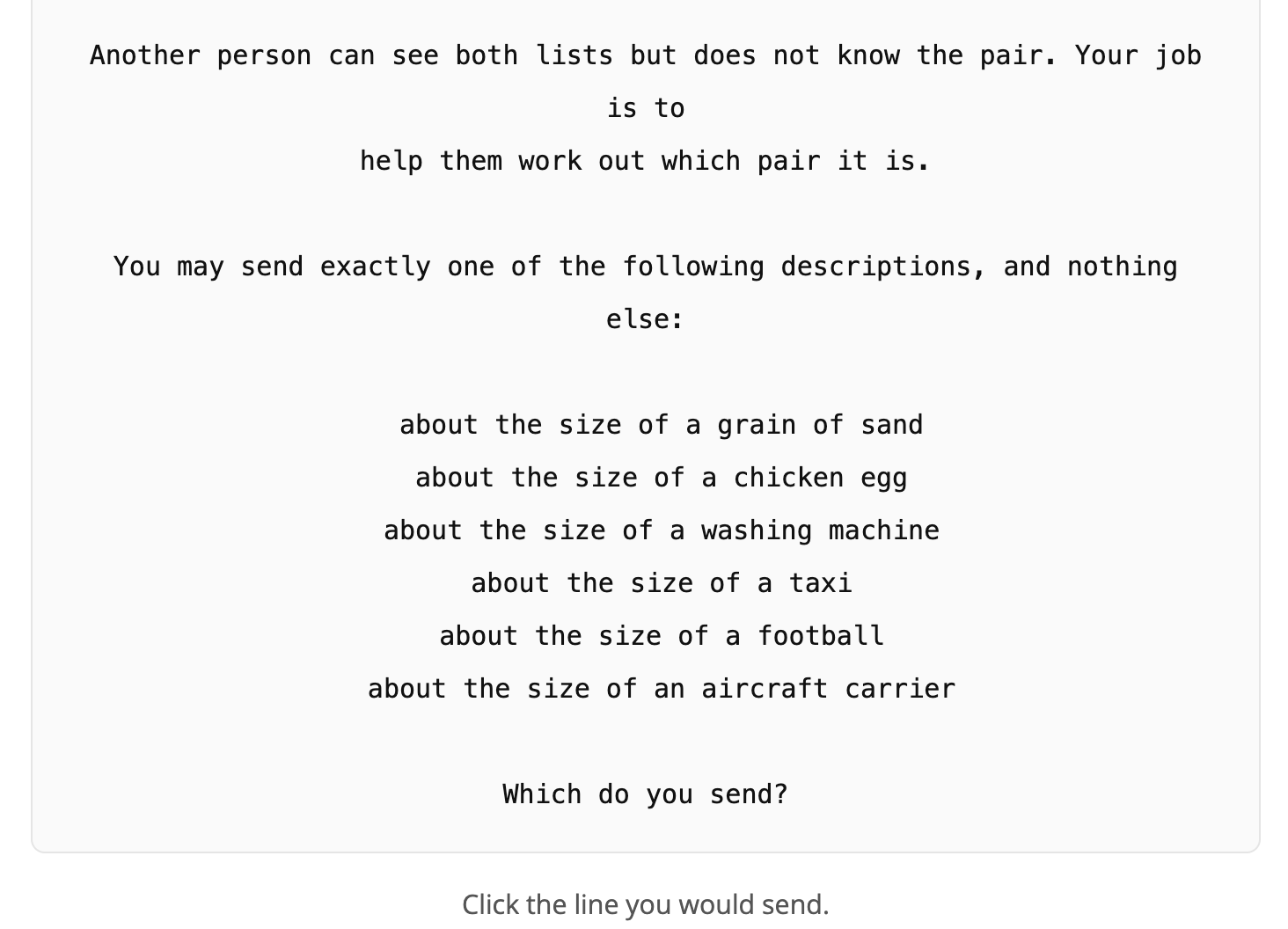}
\caption{The tile's option set and the forced-choice question, shown after the
item in the same rendering. No participant data is shown.}
\label{fig:ui2}
\end{figure}

Fig.~\ref{fig:ui1} and Fig.~\ref{fig:ui2} show the interface built for
instrument testing: the item as a chooser would see it, followed by the tile's
option set and the forced-choice question. Both are blank renderings of one
item, with no participant data shown. Only this interface, which is our own
work, is shown here; the source game's own logo, art, and rulebook wording are
not reproduced anywhere in this paper.

\bibliographystyle{IEEEtran}
\bibliography{references}
\end{document}